\documentclass{article}

\usepackage{PRIMEarxiv}
\usepackage{amsmath}
\usepackage[utf8]{inputenc} 
\usepackage[T1]{fontenc}    
\usepackage{hyperref}       
\usepackage{url}            
\usepackage{booktabs}       
\usepackage{amsfonts}       
\usepackage{nicefrac}       
\usepackage{microtype}      
\usepackage{lipsum}
\usepackage{fancyhdr}       
\usepackage{graphicx}       
\graphicspath{{media/}}     
\usepackage{natbib}
\title{CrossVIT-augmented Geospatial-Intelligence Visualization System for Tracking Economic Development Dynamics
}

\author{
  Yanbing Bai \footnotemark[1] \\
  Center for Applied Statistics and School of Statistics  \\
  Renmin University of China \\
  Beijing China\\
  \texttt{ybbai@ruc.edu.cn} \\
   \And
  Jinhua Su \footnotemark[1]\\
  Center for Applied Statistics and School of Statistics \\
  Renmin University of China \\
  Beijing China\\
  \texttt{chasesu@ruc.edu.cn} \\
   \And
  Bin Qiao \\
  School of Information \\
  Renmin University of China \\
  Beijing China\\
  \texttt{qiaobin2023@gmail.com} \\
   \And
  Xiaoran Ma \footnotemark[2]\\
  Department of Statistics and Applied Probability \\
  University of California Santa Barbara \\
  Santa Barbara US\\
  \texttt{xiaoran\_ma@ucsb.edu} \\
}

\begin{document}
\maketitle
\def\thefootnote{*}\footnotetext{These authors contributed equally to this work.}
\def\thefootnote{†}\footnotetext{Corresponding author.}

\begin{abstract}
Timely and accurate economic data is crucial for effective policymaking. Current challenges in data timeliness and spatial resolution can be addressed with advancements in multimodal sensing and distributed computing. We introduce Senseconomic, a scalable system for tracking economic dynamics via multimodal imagery and deep learning. Built on the Transformer framework, it integrates remote sensing and street view images using cross-attention, with nighttime light data as weak supervision. The system achieved an R-squared value of 0.8363 in county-level economic predictions and halved processing time to 23 minutes using distributed computing. Its user-friendly design includes a Vue3-based front end with Baidu maps for visualization and a Python-based back end automating tasks like image downloads and preprocessing. Senseconomic empowers policymakers and researchers with efficient tools for resource allocation and economic planning. The code used in this paper can be found in \href{https://github.com/HubDaniel/Satellite-Imagery-Driven-Geospatial-Intelligence-System-for-Tracking-Economic-Dynamics/tree/main}{Github}.
\end{abstract}

\keywords{Streetview Imagery \and Satellite Imagery \and Economic Dynamic Visualization \and Cross Vision transformer \and Spark Distributed Computing}

\section{Introduction}
Different levels of sensing imagery has been widely used in various human activities including agriculture \citep{yang_using_2013, nguyen_monitoring_2020}, environmental study \citep{chen_use_1997, albert_using_2017, bochenek_monitoring_2018, suel_multimodal_2021}, transportation \citep{kiwon_lee_application_2003, shaker_using_2010, abdelraouf_using_2022}, etc. The imagery contains both spatial and temporal information, providing scientists with enormous information unavailable in traditional census and economics data. Traditional approaches to tracking economic development have relied on statistical data, which often lacks spatial and temporal resolution \citep{fobi_predicting_2022, suel_multimodal_2021}. However, the rapid development of Earth observation technology has enabled the acquisition of high-resolution and timely satellite remote sensing data and street view data, providing an alternative and complementary source of geospatial data for tracking regional development patterns \citep{yeh_using_2020, rolf_generalizable_2021}. In recent years, advances in artificial intelligence (AI), remote sensing technologies, and computer vision have shown great potential in transforming the analysis of economic development using satellite imagery \citep{han_learning_2020, hall_satellite_2022}, highlighting the need for continued exploration and development of novel methodologies and techniques including integrating different sources of imagery.

Recent studies have demonstrated the utilization of satellite imagery for scoring economic development and socioeconomic status \citep{khachiyan_using_2022,abitbol_interpretable_2020,han_learning_2020}, market returns \citep{yu_eye_2023}, predicting household electricity consumption \citep{fobi_predicting_2022}, understanding economic well-being in Africa \citep{yeh_using_2020}, and mapping poverty \citep{ayush_efficient_2021, CGPM_2023}. These works have proposed deep learning-based frameworks for learning to score economic development from satellite imagery, leveraging imagery and statistical data \citep{akbari_vatt:_2021,suel_multimodal_2021}, and incorporating macroscopic social network mining \citep{CGPM_2023}. Furthermore, research has shown the potential for extracting socioeconomic indicators from high-resolution satellite images, such as urban patterns \citep{abitbol_interpretable_2020}, livelihood impact of electricity access \citep{ratledge_using_2022}, and aggregating multi-level geospatial information \citep{park_learning_2022}. Apart from solid AI models, an integrated query-extraction-visualization system is needed for better visual interpretations. A comprehensive online system opens opportunities for more users and researchers to ease their understanding and needs for scientific research.

Drawing inspiration from these studies, we integrate the latest advancements in multi-source imagery economic monitoring, supervised image processing techniques, and big data processing and visualization systems to develop a comprehensive framework for extracting and analyzing socioeconomic indicators from high-resolution satellite images. This research mainly involves the following two issues: (1) Currently, research using a single modal imagery source is prevalent \citep{han_learning_2020,yeh_using_2020,engstrom_poverty_2021,doll_mapping_2006,sutton_global_2002}, mainly focused on satellite imagery, with less use of cost-effective data sources such as street view images. There are few studies that utilize multimodal imagery sources to assess regional socio-economic indicators. Researching regional socio-economic indicators with multimodal imagery sources requires addressing the statistical consistency between street view images and satellite imagery, as well as the fusion of different modal data. (2) Concerning model-predicted economic indicators, the challenge is how to visualize them efficiently, quickly, and at low cost at the county level to aid in analysis, decision-making, and research.

Based on these two issues, considering the close relationship between nighttime light data and regional socio-economic indicators \citep{PEREZSINDIN2021100647,liu_erratum:_2021,proville_night-time_2017,gibson_which_2021,hassan_financial_2011}, this study plans to use nighttime light data as a proxy variable for local economic indicators. Using China's satellite imagery and street view images as multimodal data sources, the study aims to build a multimodal deep learning model for economic indicators and visualize the predicted results. 

In this paper, we introduce Senseconomic, an online satellite imagery-driven geospatial intelligence system for tracking economic dynamics. The system includes a systematic process to acquire, process, and analyze high-resolution satellite and street view imagery, employing a rigorous methodology for data acquisition, processing, and cropping \citep{rolf_generalizable_2021,ayush_efficient_2021}. In addition, we incorporate advancements in big data processing and visualization systems to enable the efficient analysis and presentation of results \citep{park_learning_2022,CGPM_2023}. Considering the correlation between nighttime light data and socioeconomic indicators \citep{hassan_financial_2011,proville_night-time_2017,PEREZSINDIN2021100647,liu_erratum:_2021,gibson_which_2021}, we use nighttime light data as a proxy for economic metrics. Using China’s satellite and street view imagery, this study develops a multimodal deep learning model to predict and visualize regional economic indicators effectively.

\section{Prediction Model}
In this section, we provide the architecture of the prediction model used for analyzing multi-source imagery. The model integrates techniques from both computer vision and deep learning, leveraging the Vision Transformer (ViT) and Cross-Attention mechanisms. 

\subsection{Vision Transformer} \label{vit_model}
\begin{figure}[h]  
  \centering
  \includegraphics[scale=0.4]{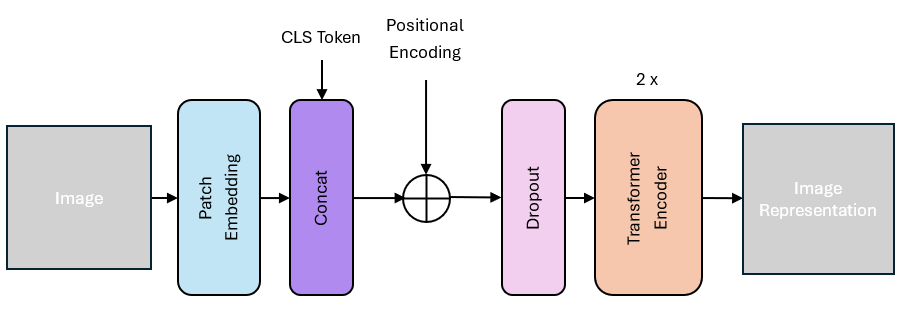}
  \caption{The structure of ViT.}
  \label{vit}
\end{figure}

In this paper, we use Vision Transformer (ViT) to extract information from images. ViT splits an image into fixed-size patches, encoding each patch into a vector using a trainable linear projection, and then processing these vectors sequentially through a transformer encoder. This encoder consists of alternating layers of multi-headed self-attention and feed-forward neural networks. The structure of ViT in this article is shown in figure \ref{vit}.

In recent years, some studies have applied satellite imagery and street view photos in the ViT model \citep{horvath_manipulation_2021,abdelraouf_using_2022} and has shown that the ViT model performs better than traditional neural networks when dealing with big data. The performance of ViT in regional socio-economic indicator assessment is promising which makes it a good candidate for extracting visual features.

\subsection{Cross-Attention} \label{cross_model}
\begin{figure}[h]  
  \centering
  \includegraphics[scale=0.4]{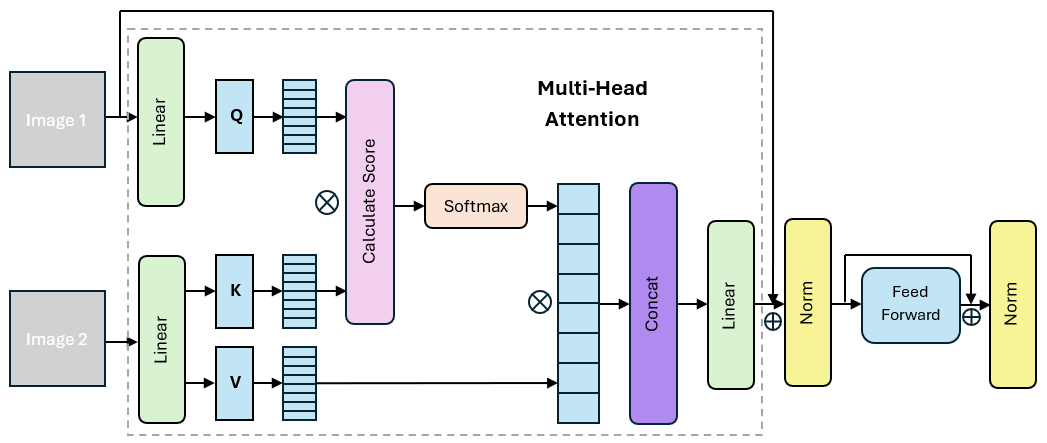}
  \caption{The structure of Cross-Attention.}
  \label{multihead}
\end{figure}

The Cross-Attention module described in the paper originates from the attention mechanism used in CrossViT \citep{chen_crossvit:_2021} for multi-scale feature fusion. It effectively captures the interconnections between different modal inputs. To adequately fuse the input data (satellite and street view imagery), the paper adopts an alternating fusion approach, performing two Cross-Attention operations on satellite imagery and street view pictures to obtain the final fused representation. Figure \ref{multihead} provides the structure of cross-attention used in this paper. 

The above Cross-Attention is run twice with 8 multi-head attention. Each head can be expressed using the following equations,

\begin{align}
    CrossAtt(image1, image2) &= image1_{1} \\
    CrossAtt(image2, image1_{1}) &= image2_{1}
\end{align}

where $image1$ and $image2$ are the representations of satellite and street view imagery from ViT. The output of the above fusion operations is the final representation of both images. It is then mapped to a scalar using a fitting head which includes a linear layer, a ReLU activation, and another linear layer as the prediction of nighttime light data.

\section{Datasets}
To meet the requirements of model training and the visualization system, this paper utilizes data from multiple sources. The data for model training includes remote sensing imagery, street view images, and nighttime light data, while the data for the visualization system includes remote sensing imagery, street view images, administrative division data, and statistical yearbook data. Higher-quality remote sensing imagery from the ArcGIS website was used for model training, whereas the system used imagery data from the Sentinel-2 remote sensing satellite for visualization. The following provides a detailed description of the data sources.

\subsection{Data for Model Training}
\textbf{Satellite Imagery.} 
Satellite Imagery comes from the ArcGIS website, an internet-based Geographic Information System (GIS) platform provided by Esri. The website provides global satellite remote sensing imagery accessible to users. This imagery is structured in a pyramidal format, tiered across 19 levels (also known as zoom levels) to accommodate various resolution requirements. We selected zoom level 12, at which the resolution of the satellite imagery is approximately 38.4m. Each image file is 256 pixels by 256 pixels, covering an area of about 9.83km by 9.83km. This resolution retains rich ground details suitable for studying regional economic development. Within the national scope, a total of 123,758 remote-sensing images were collected. 

\textbf{Street Imagery.} 
Street Imagery comes from the Baidu Maps Panorama Platform. Baidu Maps' panorama service provides a 360° view of the real world, enabling users to browse street views from different angles by dragging the map. The most common and clear street view image parameters are used. 

\begin{figure}[h]  
  \centering
  \includegraphics[scale=0.6]{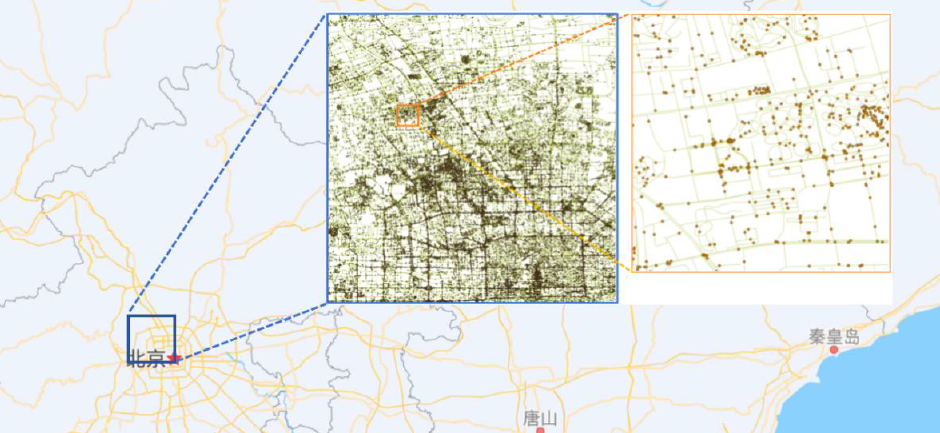}
  \caption{Example of sampled street view locations in Beijing.}
  \label{street1}
\end{figure}

A total of 112,321 street-view images covering the national range were obtained, with a vertical angle of 0 degrees, horizontal angles at 0, 90, 180, and 270 degrees, and an image size of 480 pixels by 320 pixels with a horizontal field of view of 90 degrees. Figure \ref{street1} shows randomly generated street views in some urban areas of Beijing. Our method obtains street viewpoints in a relatively uniform manner, with an average sampling interval of about 110m. 

\textbf{Nighttime Light Data.} 
The data comes from Suomi National Polar-orbiting Partnership (NPP) satellite, launched by NASA in 2011. It is designed to collect data on the Earth's atmosphere, clouds, oceans, land, and ice and snow. The Day/Night Band (DNB) of VIIRS provides high-resolution nighttime light data, covering the globe daily. Stray light is a common interference factor that can affect the accuracy and reliability of the data. Stray light, caused by the scattering and reflection of nighttime light radiation, can lead to anomalies in brightness or the spread of light spots in remote sensing images. The NPP/VIIRS provides the VCMSL (Visible Infrared Imaging Radiometer Suite Cloud Mask Stray Light) tool to address the effects of stray light. The data used in this paper are the VCMSL data products from 2020, corrected for stray light. The format is raster type, with a spatial resolution of approximately $742m*742m$. The data's geographic coordinates range from longitude $-180.00208333335$ to $180.00208621335$ and latitude from $-65.00208445335001$ to $75.00208333335$. The difference in latitude and longitude between adjacent data points is $0.0041666667$ degrees.

\subsection{Data for System}
\begin{figure}[h]  
  \centering
  \includegraphics[scale=0.4]{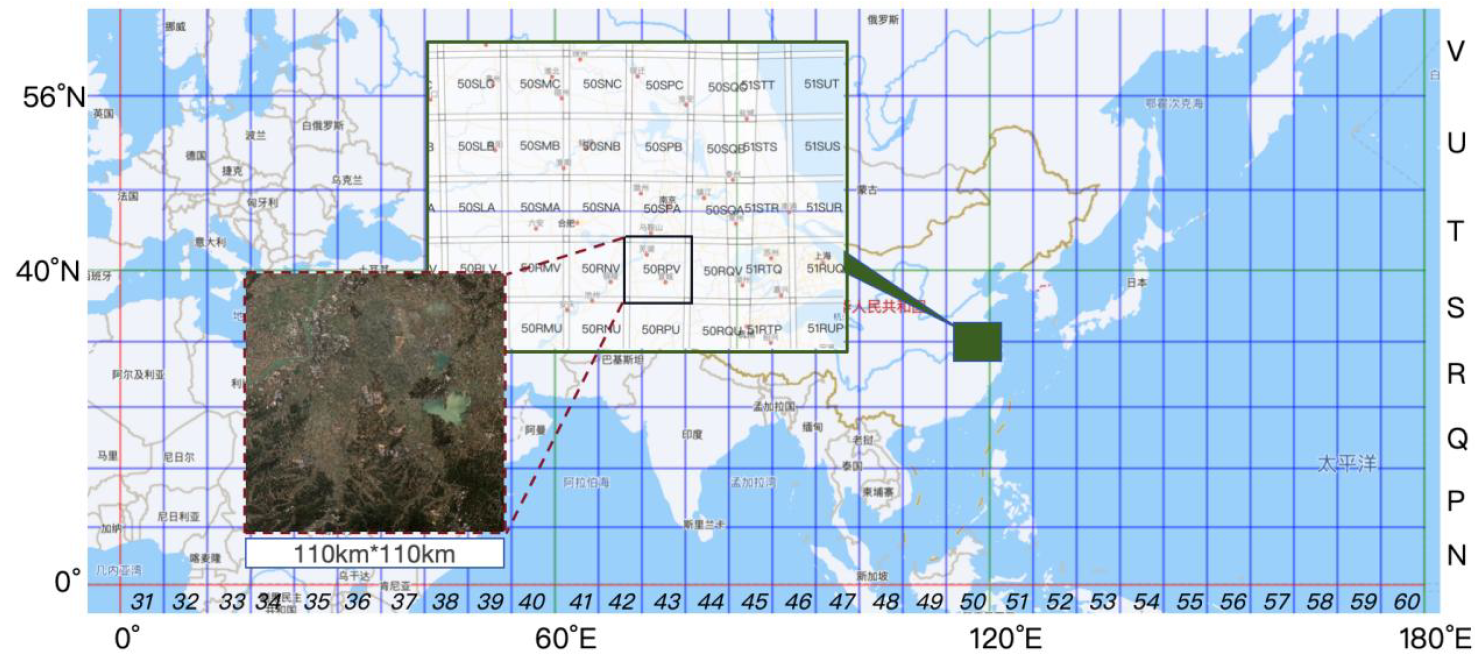}
  \caption{Example of Level-2A satellite images.}
  \label{satellite_image}
\end{figure}

\textbf{Remote Sensing Data.}
The Sentinel-2 environmental monitoring satellite is an Earth observation satellite launched by the European Space Agency (ESA) in 2015, aimed at providing high-resolution, multispectral surface imagery for global environmental monitoring and sustainable development. The provided products have fixed-size granules (also known as tiles), divided into multiple levels. Typically, ordinary users utilize Level-1C and Level-2A data, both adopting $110km \times 110km$ image granules. Figure \ref{satellite_image} is an example of the satellite data. Each square represents a predefined tile by the product, and each tile has a surface area of $110 \times 110 km^2$ to ensure substantial overlap with adjacent tiles. The primary difference between Level-1C and Level-2A is that Level-2A is atmospherically corrected, reducing the inaccuracies in surface reflectance caused by atmospheric interference. This paper uses Level-2A data which is obtained through API calls.

\textbf{Street View Data.}
The street view data used for the system are sourced and acquired in the same manner as those used for model training.

\textbf{Administrative Division Data.}
The system's visualization base map utilizes Baidu Maps, customizing the map style through the platform's personalized map customization feature. The final economic indicators need to be aggregated to areas defined at the county-level administrative division, thus requiring boundary coordinate data of administrative divisions. Baidu Maps' API can obtain province, city, and county names and the coordinate sequence of the area based on latitude and longitude values, collecting administrative division data by traversing the target area's latitude and longitude and entering it into the basic database. The county area data format is a sequence of coordinate pairs separated by commas.

\textbf{County-Level Statistical Yearbook Data.}
The system uses GDP economic data from the "China County Statistical Yearbook" for a comparative analysis of 110 target counties. The yearbook includes basic information and comprehensive economic data of more than two thousand county-level units across the country, providing a comprehensive reflection of China's county-level socio-economic development. The system utilizes economic data from 2019 and 2020. For counties with missing data, annual GDP statistical data published on the official county websites are searched to complete the records.

\subsection{Data Processing}
\subsubsection{Data Alignment} \label{data_alignment}
\begin{figure}[h]  
  \centering
  \includegraphics[scale=0.6]{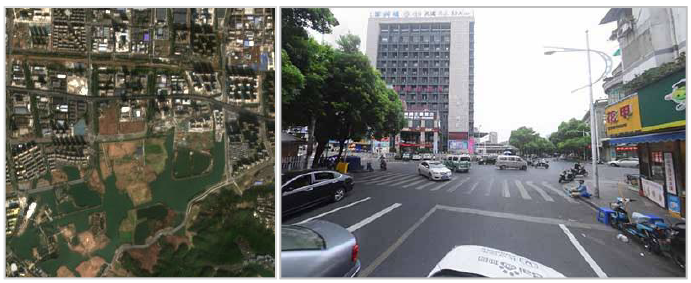}
  \caption{A satellite-street view pair.}
  \label{pairimage}
\end{figure}

This paper employs a data alignment method based on satellite imagery, with satellite images as the primary data and street view pictures as supplementary data. The specific approach involves calculating the distance between the latitude and longitude labels of both satellite and street view images. For each satellite image, the closest street view picture is selected to form a satellite-street view pair. The study generated 124,758 such image-street data pairs, with varying distances between them, some quite far apart. Given the resolution of the economic activity map selected for this study is $5km$ (using the average nighttime light value within a $5km \times 5km$ area), the pairs where both elements are within $5km$ of each other are further filtered based on the selection criteria of the nighttime light index. Ultimately, 38,126 data pairs were obtained for model training and testing, covering 860 counties nationwide. Figure \ref{pairimage} illustrates a satellite-street view pair, clearly showing the differences in how satellite data and street view data reflect socio-economic information of the same area: satellite imagery focuses on the regional overview, highlighting information such as the condition of buildings, road width, and vegetation coverage; street view data focuses on local information, mainly reflecting fine visual details like houses, cars, signs, and shops around the area’s central point.

\subsubsection{Data Preprocessing} \label{data_proc}

Both satellite images and street view pictures underwent the following preprocessing methods.

\textbf{Image Resizing.}
Images were resized to $224 \times 224$. This image size allows for fine-tuning under multiple pre-trained models. For instance, the classic model ResNet-18 is designed for $224 \times 224$ images.

\textbf{Z-Normalization.}
Through z-normalization, data are transformed to have a distribution with a mean close to 0 and a variance close to 1. This enhances data stability, mitigates the impact of outliers, better represents data characteristics, and improves model convergence speed and stability, thereby enhancing training efficiency.

\textbf{Data Augmentation.}
In situations of limited data volume, random flips, rotations, and cropping can effectively increase the number of samples, enhancing the diversity of the training set, reducing the risk of model overfitting, and improving the model’s generalization ability. Especially with remote sensing imagery, random flips and rotations can simulate different observational angles and perspectives of satellites, enhancing model performance against various angles and directions, reducing dependency on specific data, and improving robustness.

\section{System Architecture}
\begin{figure}[h]  
  \centering
  \includegraphics[scale=0.6]{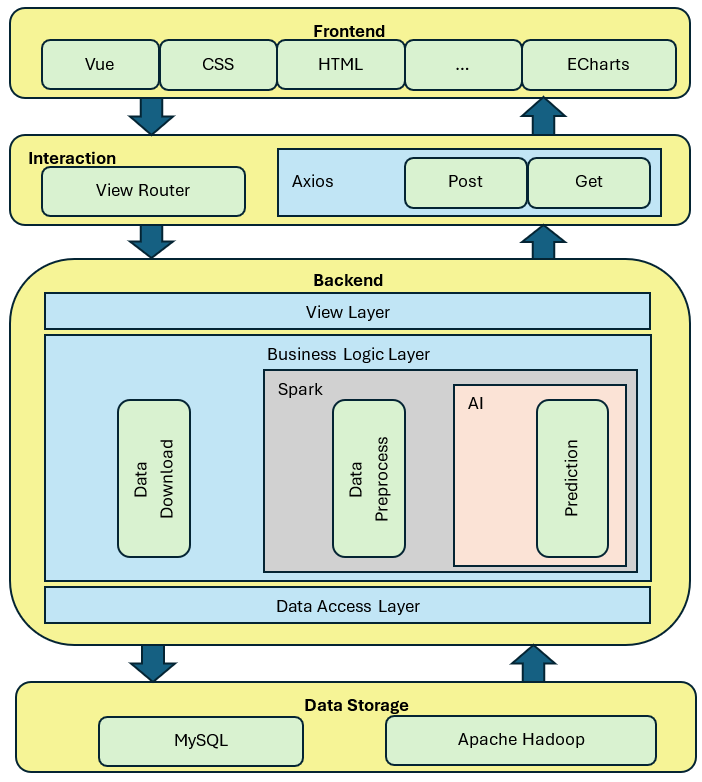}
  \caption{The structure of system architecture.}
  \label{system}
\end{figure}

To better visualize the prediction results of the model and assist in decision-making and research work, Senseconomic, a human-computer interaction system that meets specific needs is developed, guided by software engineering theory, and combined with Vue frontend visualization technology and data management techniques. This system implements a dynamic economic visualization system at the county-level administrative region granularity. The system architecture is shown in figure \ref{system}.

The system adopts a B/S (Browser/Server) model, using Python as the server-side development language, MySQL as the foundational database, Hadoop as the distributed system framework, and Spark as the distributed computing engine. It integrates the JavaScript API GL service of Baidu Maps and uses Visual Studio Code as the development environment to implement data collection, computation, storage, and visualization.

The server framework uses the Django Web framework, deployed using Nginx in conjunction with uWSGI. uWSGI acts as the Python web server, using the WSGI (Web Server Gateway Interface) protocol to interact with Django projects and the uwsgi protocol to communicate with Nginx. The client side primarily uses HTML, CSS, and JavaScript technologies, combined with the Vue framework for construction. Vue is a simple, efficient frontend framework that facilitates rapid user interface construction, offering a component-based programming model while retaining HTML's declarative programming basis and enhancing code maintainability. The pages utilize the View UI Plus component library, which provides a wide range of components and functionalities suitable for most website scenarios. Data analysis charts are rendered using Apache ECharts, an open-source visualization chart library based on JavaScript.

The system is developed with a frontend-backend separation approach, and the interface design conforms to the REST (Representational State Transfer) style. The front end accesses interfaces through Axios, a Promise-based HTTP client for the browser and Node.js. Following the design principle of high cohesion and low coupling, the backend services are structured in a three-tier architecture: the data access layer, the business logic layer, and the view layer. The data access layer is responsible for data maintenance and persistence. The business logic layer handles the implementation of various business logic, while the view layer processes requests from users, determining which business module to engage.

\section{Distributed Computing}
Spark \citep{han_spark:_2015} is a distributed computing system designed to handle large-scale data sets and provide fast and powerful distributed computing capabilities. It utilizes in-memory computing and the Resilient Distributed Dataset (RDD) model, allowing for efficient data processing tasks on large-scale clusters.

In our system, a large amount of images can be processed in a parallel fashion for both image preprocessing and prediction. Tasks are submitted through the command line using os.system and spark-submit to execute Python files with parameters retrieved via sys.argv. The core computation for image processing involves 

\begin{enumerate}
    \item Environment Configuration: Setting up the environment with SparkConf, specifying master and job name.
    \item Data Reading: Retrieving and reading directory addresses for remote sensing images and street views using Spark's methods to create RDDs.
    \item Data Processing: Using map and join operations to align (more in section \ref{data_alignment}) remote sensing images with street views and calculate scores. These scores are grouped and processed to determine maximum values per index.
    \item Result Handling: The computation stages are visualized in a DAG, divided into three stages focusing on file reading, score prediction, and maximum score selection. The results are temporarily saved to Hadoop and then transferred to a database for persistent storage.
\end{enumerate}

For model predictions, the models are loaded with configurations from a .yaml file, instantiated, and weights are loaded. Then the images undergo transformations such as resizing and normalization to prepare for prediction. The model uses these processed images to output prediction scores. Finally, data after alignment and prediction are saved and used to update the database, contributing to the system’s ongoing analysis and output generation.

\section{Database Design}
The data storage in this paper uses the relational database MySQL, utilizing street view data and remote sensing image data to compute economic indicators at the county-level granularity. The database objects in this research primarily include tables for remote sensing image frame information, street viewpoint data, administrative region information, economic prediction tasks, remote sensing image frame processing, fine-grained remote sensing image information, county-level economic prediction results, and user information. In forecasting work for different regions and times, the system manages tasks through the economic prediction task table, which stores task information. The remote sensing image frame processing table records specific processing task progress for each image in the target area. A large remote sensing image is processed into many small images, which are then used in conjunction with street view images; the fine-grained remote sensing image information table records information about these small images and the fine-grained prediction scores, which are considered intermediate results and can be seen as a temporary table. The county-level economic prediction results table records the final consolidated results at the county administrative level, which can be used to provide query services externally. This paper employs the E-R diagram method for the design of the data entity model. It follows the three normal forms for logical database design, using the most basic data units to create database tables, and setting unique primary keys to ensure direct relationships exist between the primary key and non-primary attributes.

\section{Frontend Design}
\begin{figure}[h]  
  \centering
  \includegraphics[scale=0.6]{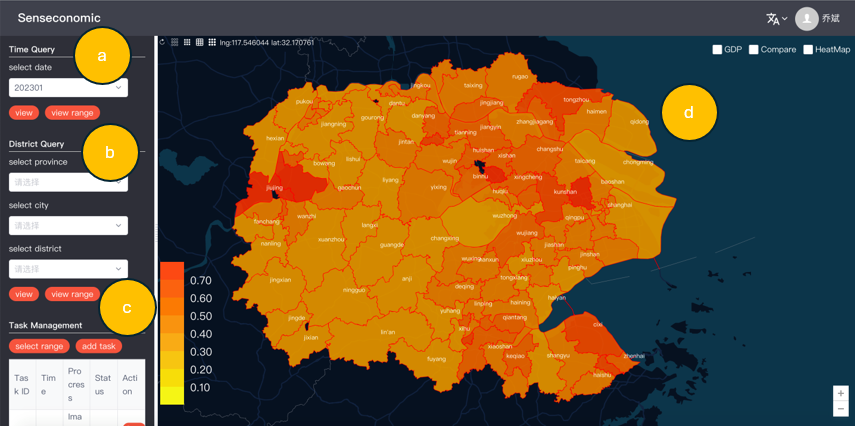}
  \caption{Screenshot of the system UI. (a) Time selection and viewing. (b) The data processing process operation panel: administrative region selection. After setting parameters such as start time and end time, a new data processing process can be started. (c) Log output of data processing process. (d) On the right side is the map visualization display of processing results. }
  \label{system1}
\end{figure}

The client interface of this system is primarily built using the Vue3 framework in combination with the View UI Plus component library. Vue uses traditional templating, which helps lower transition costs compared to React's JSX, which requires additional learning. Vue is a progressive framework that allows for gradual integration during development, offering great flexibility and a smaller size—only half that of React, and it performs better in terms of speed. Vue also comes with official support for common web application development tools like routing, state management, unit testing, and static site generation, which contribute to a better ecosystem. Therefore, the front end of this system is developed using Vue. An illustration of the system UI is shown in Figure \ref{system1}.

\section{Application}
\subsection{Feature Extraction and Fusion}
In this paper, the feature extraction from image data is done using the ViT model described in section \ref{vit_model}. The input data are preprocessed images (section \ref{data_proc}) with dimensions $3 \times 224 \times 224$. The traditional ViT patch-embedding method is slightly modified to accommodate a representation with a hidden dimension of 256. Specifically, each $224 \times 224$ image is divided into 49 patches of $32 \times 32$. Each patch is then flattened into a vector, resulting in a $49 \times 3072$ dimensional vector. This is subsequently compressed through a sequence of Layer Normalization and Linear layers to a $49 \times 256$ dimensional vector.

In this study, a CLS token is used as a prediction token with a dimension of $1 \times 256$, which is randomly initialized and concatenated to the front of the Patch-Embedding output. During prediction, the transformed value of this token is extracted for further prediction. This study uses two layers of Transformer-Encoders. The input and output of the Transformer-Encoder are both $50 \times 256$. The output from the second layer represents the learned image representation.

The cross-attention mechanism introduced in section \ref{cross_model} is used for feature fusion.

\subsection{Prediction Model Performance}
The aligned multimodal dataset is randomly split into two parts, with 80\% used for model training and the remaining 20\% used for model validation. The dropout rate for all Dropout layers is 0.2. The batch size for training data is set to 256, with 40 epochs of iterations. The MSE loss and the Adam gradient descent method are used for optimization, with an initial learning rate of 0.0001. The model is trained and evaluated on a single Nvidia-V100-GPU using the PyTorch deep learning framework.

We run our model under different settings to evaluate its performance. ResNet-18 is used as an alternative to ViT-B16 for feature extraction and we also checked the difference between using unimodal and multi-modal data. The R-squared (coefficient of determination) is used as the model evaluation metric, with a range from 0 to 1, where closer to 1 indicates a better model fit. The formula for R-squared is as follows:

$$
R^2 = 1 - \frac{\sum_{i=1}^{n} (\hat{y}_i - y_i)^2}{\sum_{i=1}^{n} (\bar{y} - y_i)^2}
$$

\begin{table}[h]
\caption{Coefficient of determination for different prediction results.}
\label{Tab:result}
\centering
\begin{tabular}{ cccc } 
\toprule 
Data & Model & Fusion Method & $R^2$ \\
\midrule
Satellite & ResNet-18 & - & 0.8318\\
Street View & ResNet-18 & - & 0.2610\\
Satellite & ViT-B16 & - & 0.7801\\
Street View & ViT-B16 & - & 0.2262\\
Satellite \& Street View & ViT-B16 & CrossAtt & \textbf{0.8363}\\
\bottomrule
\end{tabular}
\end{table}

Table \ref{Tab:result} lists the R-squared results obtained on the test set. When fitting regional socio-economic indicators as proxy variables, the fitting ability of remote sensing satellite data is far stronger than street view data because the satellite imagery provides extensive information on farmlands, buildings, roads, and other land features, which reflect the level of economic development and economic activity in the region. The fitting effect of ResNet-18 is slightly better than that of ViT-B16, but with more data and complex model structure in future versions, the ViT model could have a better performance. The multimodal model using both satellite and street view data performs better than the unimodal model using satellite data alone, suggesting that the introduction of street view data could improve performance.

\begin{figure}[h]  
  \centering
  \includegraphics[scale=0.6]{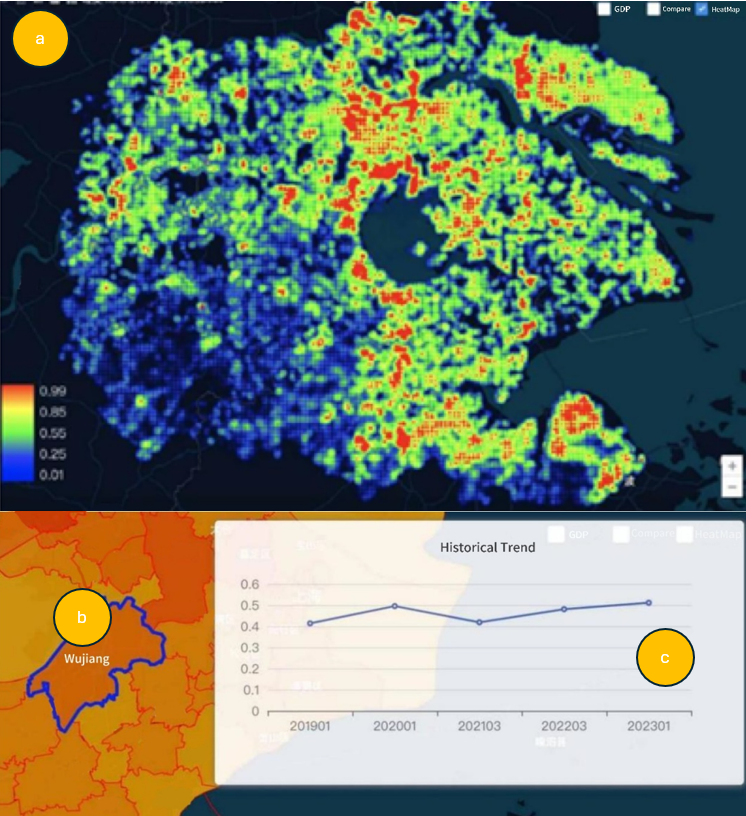}
  \caption{The heat map (a) of the region in 2020 is generated based on the predicted scores and the center point of the grid, with high scores in red and low scores in green. (b) is the detailed area of Wujiang District, Suzhou City, Jiangsu Province. (c) is the trend chart of predicted value from 2019 to 2023.}
  \label{system2}
\end{figure}

An illustration of the prediction results and its visualization can be found in Figure \ref{system2} panel a. The system also allows for viewing the fine-grained data of a particular county and its historical trend as shown in Figure \ref{system2} panel b and c).

\subsection{System Performance}
The experimental environment used in this paper consists of a Spark cluster with two servers, where one server is deployed as both Master and Worker, and the other as a Worker. The basic information of the server nodes and their dependent tools are listed in Table \ref{Tab:conf_inf}.

\begin{table}[h]
\caption{Server configuration information.}
\centering
\begin{tabular}{ cc } 
    \toprule
Name & Configuration \\
    \midrule
CPU Information & Intel(R) Core(TM) i5-8257U CPU \@ 1.40GHz\\
CPU Cores & 6\\
Memory & 8G\\
Disk & 100G\\
Operating System & CentOS-7-x86\_64-Minimal-2207-02\\
JDK & jdk-8u381-linux-x64\\
Hadoop & hadoop-3.3.6\\
Spark & spark-3.4.1-bin-hadoop3\\
  \bottomrule
\end{tabular}
\label{Tab:conf_inf}
\end{table}

During the data publishing process of the system, the bottlenecks in performance occur during the image download and model prediction phases. Image downloading is significantly affected by the network environment, making the model prediction phase a critical factor affecting system performance. Under the server environment described, experiments were conducted in both Spark cluster mode and non-Spark mode (without Spark and Hadoop environment, using a single server) on remote sensing images numbered 51STR and 51RTQ from January 2023. These images correspond to 14,682 street-view pictures. The running times are shown in Table \ref{Tab:perform}, which demonstrates that Spark significantly improves execution efficiency.

\begin{table}[h]
\caption{Experimental results for 51STR images.}
\centering
\begin{tabular}{ ccccc } 
\toprule 
Frame & Num. Satelite & Num. Street View & Mode & Time \\
\midrule
51STR & 1600 & 14682 & Spark & 19min\\
51STR & 1600 & 14682 & non-Spark & 42min\\
51RTQ & 1600 & 17131 & Spark & 23min\\
51RTQ & 1600 & 17131 & non-Spark & 49min\\
\bottomrule
\end{tabular}
\label{Tab:perform}
\end{table}

\section{Conclusion}
This paper proposes, Senseconomic, a geospatial intelligence system for tracking economic dynamics using multi-source imagery. The system uses a multimodal deep learning framework based on Transformer and cross attention, incorporating information from remote sensing image data and street view images. The nighttime light data is used as a proxy variable for socio-economic indices for training the prediction model and evaluating regional socio-economic indicators. The paper focuses on using this model, combined with Spark distributed computing and frontend visualization technologies, to design and implement a system for visualizing regional economic dynamics, providing visual support for government and business decision-making. 

This study contributes to the growing body of research in geospatial intelligence
and multimodal imagery analysis, with implications for policymakers, urban
planners, and researchers interested in understanding the complex interplay between
economic development and geographic factors. 

With promising results achieved so far, however, the authors realize the potential for improvement. Future work will focus on enhancing the model's accuracy and efficiency, exploring additional data sources, and refining the system's scalability and user interface. These improvements aim to make Senseconomic an even more powerful tool for economic analysis and decision-making.


\bibliographystyle{unsrtnat}  
\bibliography{reference}

\begin{thebibliography}{33}
\providecommand{\natexlab}[1]{#1}
\providecommand{\url}[1]{\texttt{#1}}
\expandafter\ifx\csname urlstyle\endcsname\relax
  \providecommand{\doi}[1]{doi: #1}\else
  \providecommand{\doi}{doi: \begingroup \urlstyle{rm}\Url}\fi

\bibitem[Yang et~al.(2013)Yang, Everitt, Du, Luo, and Chanussot]{yang_using_2013}
Chenghai Yang, James~H. Everitt, Qian Du, Bin Luo, and Jocelyn Chanussot.
\newblock Using high-resolution airborne and satellite imagery to assess crop growth and yield variability for precision agriculture.
\newblock \emph{Proceedings of the IEEE}, 101\penalty0 (3):\penalty0 582--592, March 2013.
\newblock ISSN 0018-9219, 1558-2256.
\newblock \doi{10.1109/JPROC.2012.2196249}.
\newblock URL \url{http://ieeexplore.ieee.org/document/6236221/}.

\bibitem[Nguyen et~al.(2020)Nguyen, Hoang, Pham, Vu, Nguyen, Huynh, and Jo]{nguyen_monitoring_2020}
Thanh~Tam Nguyen, Thanh~Dat Hoang, Minh~Tam Pham, Tuyet~Trinh Vu, Thanh~Hung Nguyen, {Quyet-Thang} Huynh, and Jun Jo.
\newblock Monitoring agriculture areas with satellite images and deep learning.
\newblock \emph{Applied Soft Computing}, 95:\penalty0 106565, October 2020.
\newblock ISSN 15684946.
\newblock \doi{10.1016/j.asoc.2020.106565}.
\newblock URL \url{https://linkinghub.elsevier.com/retrieve/pii/S1568494620305032}.

\bibitem[Chen et~al.(1997)Chen, Chen, Chang, and Chen]{chen_use_1997}
C.F. Chen, K.S. Chen, L.Y. Chang, and A.J. Chen.
\newblock The use of satellite imagery for monitoring coastal environment in {Taiwan}.
\newblock In \emph{{IGARSS}'97. 1997 {IEEE} {International} {Geoscience} and {Remote} {Sensing} {Symposium} {Proceedings}. {Remote} {Sensing} - {A} {Scientific} {Vision} for {Sustainable} {Development}}, volume~3, pages 1424--1426, Singapore, 1997. IEEE.
\newblock ISBN 9780780338364.
\newblock \doi{10.1109/IGARSS.1997.606466}.
\newblock URL \url{http://ieeexplore.ieee.org/document/606466/}.

\bibitem[Albert et~al.(2017)Albert, Kaur, and Gonzalez]{albert_using_2017}
Adrian Albert, Jasleen Kaur, and Marta~C. Gonzalez.
\newblock Using convolutional networks and satellite imagery to identify patterns in urban environments at a large scale.
\newblock In \emph{Proceedings of the 23rd {ACM} {SIGKDD} {International} {Conference} on {Knowledge} {Discovery} and {Data} {Mining}}, pages 1357--1366, Halifax NS Canada, August 2017. ACM.
\newblock ISBN 9781450348874.
\newblock \doi{10.1145/3097983.3098070}.
\newblock URL \url{https://dl.acm.org/doi/10.1145/3097983.3098070}.

\bibitem[Bochenek et~al.(2018)Bochenek, Ziolkowski, Bartold, Orlowska, and Ochtyra]{bochenek_monitoring_2018}
Zbigniew Bochenek, Dariusz Ziolkowski, Maciej Bartold, Karolina Orlowska, and Adrian Ochtyra.
\newblock Monitoring forest biodiversity and the impact of climate on forest environment using high-resolution satellite images.
\newblock \emph{European Journal of Remote Sensing}, 51\penalty0 (1):\penalty0 166--181, January 2018.
\newblock ISSN 2279-7254.
\newblock \doi{10.1080/22797254.2017.1414573}.
\newblock URL \url{https://www.tandfonline.com/doi/full/10.1080/22797254.2017.1414573}.

\bibitem[Suel et~al.(2021)Suel, Bhatt, Brauer, Flaxman, and Ezzati]{suel_multimodal_2021}
Esra Suel, Samir Bhatt, Michael Brauer, Seth Flaxman, and Majid Ezzati.
\newblock Multimodal deep learning from satellite and street-level imagery for measuring income, overcrowding, and environmental deprivation in urban areas.
\newblock \emph{Remote Sensing of Environment}, 257:\penalty0 112339, May 2021.
\newblock ISSN 00344257.
\newblock \doi{10.1016/j.rse.2021.112339}.
\newblock URL \url{https://linkinghub.elsevier.com/retrieve/pii/S0034425721000572}.

\bibitem[{Kiwon Lee} et~al.(2003){Kiwon Lee}, {Se-Kyung Oh}, and {Hee-Young Ryu}]{kiwon_lee_application_2003}
{Kiwon Lee}, {Se-Kyung Oh}, and {Hee-Young Ryu}.
\newblock Application of high-resolution satellite imagery to transportation: accessibility index extraction approach.
\newblock In \emph{{IGARSS} 2003. 2003 {IEEE} {International} {Geoscience} and {Remote} {Sensing} {Symposium}. {Proceedings} ({IEEE} {Cat}. {No}.{03CH37477})}, volume~4, pages 2942--2944, Toulouse, France, 2003. IEEE.
\newblock ISBN 9780780379299.
\newblock \doi{10.1109/IGARSS.2003.1294639}.
\newblock URL \url{http://ieeexplore.ieee.org/document/1294639/}.

\bibitem[Shaker et~al.(2010)Shaker, Yan, and Easa]{shaker_using_2010}
Ahmed Shaker, Wai~Yeung Yan, and Said Easa.
\newblock Using stereo satellite imagery for topographic and transportation applications: an accuracy assessment.
\newblock \emph{GIScience \& Remote Sensing}, 47\penalty0 (3):\penalty0 321--337, July 2010.
\newblock ISSN 1548-1603, 1943-7226.
\newblock \doi{10.2747/1548-1603.47.3.321}.
\newblock URL \url{https://www.tandfonline.com/doi/full/10.2747/1548-1603.47.3.321}.

\bibitem[Abdelraouf et~al.(2022)Abdelraouf, Abdel-Aty, and Wu]{abdelraouf_using_2022}
Amr Abdelraouf, Mohamed Abdel-Aty, and Yina Wu.
\newblock Using vision transformers for spatial-context-aware rain and road surface condition detection on freeways.
\newblock \emph{IEEE Transactions on Intelligent Transportation Systems}, 23\penalty0 (10):\penalty0 18546--18556, October 2022.
\newblock ISSN 1524-9050, 1558-0016.
\newblock \doi{10.1109/TITS.2022.3150715}.
\newblock URL \url{https://ieeexplore.ieee.org/document/9716073/}.

\bibitem[Fobi et~al.(2022)Fobi, Mugyenyi, Williams, Modi, and Taneja]{fobi_predicting_2022}
Simone Fobi, Joel Mugyenyi, Nathaniel~J. Williams, Vijay Modi, and Jay Taneja.
\newblock Predicting levels of household electricity consumption in low-access settings.
\newblock In \emph{2022 {IEEE}/{CVF} {Winter} {Conference} on {Applications} of {Computer} {Vision} ({WACV})}, pages 2213--2222, Waikoloa, HI, USA, January 2022. IEEE.
\newblock ISBN 9781665409155.
\newblock \doi{10.1109/WACV51458.2022.00227}.
\newblock URL \url{https://ieeexplore.ieee.org/document/9706653/}.

\bibitem[Yeh et~al.(2020)Yeh, Perez, Driscoll, Azzari, Tang, Lobell, Ermon, and Burke]{yeh_using_2020}
Christopher Yeh, Anthony Perez, Anne Driscoll, George Azzari, Zhongyi Tang, David Lobell, Stefano Ermon, and Marshall Burke.
\newblock Using publicly available satellite imagery and deep learning to understand economic well-being in africa.
\newblock \emph{Nature Communications}, 11\penalty0 (1):\penalty0 2583, May 2020.
\newblock ISSN 2041-1723.
\newblock \doi{10.1038/s41467-020-16185-w}.
\newblock URL \url{https://www.nature.com/articles/s41467-020-16185-w}.

\bibitem[Rolf et~al.(2021)Rolf, Proctor, Carleton, Bolliger, Shankar, Ishihara, Recht, and Hsiang]{rolf_generalizable_2021}
Esther Rolf, Jonathan Proctor, Tamma Carleton, Ian Bolliger, Vaishaal Shankar, Miyabi Ishihara, Benjamin Recht, and Solomon Hsiang.
\newblock A generalizable and accessible approach to machine learning with global satellite imagery.
\newblock \emph{Nature Communications}, 12\penalty0 (1):\penalty0 4392, July 2021.
\newblock ISSN 2041-1723.
\newblock \doi{10.1038/s41467-021-24638-z}.
\newblock URL \url{https://www.nature.com/articles/s41467-021-24638-z}.

\bibitem[Han et~al.(2020)Han, Ahn, Park, Yang, Lee, Kim, Yang, Park, and Cha]{han_learning_2020}
Sungwon Han, Donghyun Ahn, Sungwon Park, Jeasurk Yang, Susang Lee, Jihee Kim, Hyunjoo Yang, Sangyoon Park, and Meeyoung Cha.
\newblock Learning to score economic development from satellite imagery.
\newblock In \emph{Proceedings of the 26th {ACM} {SIGKDD} {International} {Conference} on {Knowledge} {Discovery} \& {Data} {Mining}}, pages 2970--2979, Virtual Event CA USA, August 2020. ACM.
\newblock ISBN 9781450379984.
\newblock \doi{10.1145/3394486.3403347}.
\newblock URL \url{https://dl.acm.org/doi/10.1145/3394486.3403347}.

\bibitem[Hall et~al.(2022)Hall, Ohlsson, and {Rögnvaldsson}]{hall_satellite_2022}
Ola Hall, Mattias Ohlsson, and Thorsteinn {Rögnvaldsson}.
\newblock Satellite image and machine learning ased knowledge extraction in the poverty and welfare domain.
\newblock \emph{SSRN Electronic Journal}, 2022.
\newblock ISSN 1556-5068.
\newblock \doi{10.2139/ssrn.4102620}.
\newblock URL \url{https://www.ssrn.com/abstract=4102620}.

\bibitem[Khachiyan et~al.(2022)Khachiyan, Thomas, Zhou, Hanson, Cloninger, Rosing, and Khandelwal]{khachiyan_using_2022}
Arman Khachiyan, Anthony Thomas, Huye Zhou, Gordon Hanson, Alex Cloninger, Tajana Rosing, and Amit~K. Khandelwal.
\newblock Using neural networks to predict microspatial economic growth.
\newblock \emph{American Economic Review: Insights}, 4\penalty0 (4):\penalty0 491--506, December 2022.
\newblock ISSN 2640-205X, 2640-2068.
\newblock \doi{10.1257/aeri.20210422}.
\newblock URL \url{https://pubs.aeaweb.org/doi/10.1257/aeri.20210422}.

\bibitem[Abitbol and Karsai(2020)]{abitbol_interpretable_2020}
Jacob~Levy Abitbol and {Márton} Karsai.
\newblock Interpretable socioeconomic status inference from aerial imagery through urban patterns.
\newblock \emph{Nature Machine Intelligence}, 2\penalty0 (11):\penalty0 684--692, October 2020.
\newblock ISSN 2522-5839.
\newblock \doi{10.1038/s42256-020-00243-5}.
\newblock URL \url{https://www.nature.com/articles/s42256-020-00243-5}.

\bibitem[Yu et~al.(2023)Yu, Hao, Wu, Zhao, and Wang]{yu_eye_2023}
Honghai Yu, Xianfeng Hao, Liangyu Wu, Yuqi Zhao, and Yudong Wang.
\newblock Eye in outer space: satellite imageries of container ports can predict world stock returns.
\newblock \emph{Humanities and Social Sciences Communications}, 10\penalty0 (1):\penalty0 383, July 2023.
\newblock ISSN 2662-9992.
\newblock \doi{10.1057/s41599-023-01891-9}.
\newblock URL \url{https://www.nature.com/articles/s41599-023-01891-9}.

\bibitem[Ayush et~al.(2021)Ayush, Uzkent, Tanmay, Burke, Lobell, and Ermon]{ayush_efficient_2021}
Kumar Ayush, Burak Uzkent, Kumar Tanmay, Marshall Burke, David Lobell, and Stefano Ermon.
\newblock Efficient poverty mapping from high resolution remote sensing images.
\newblock \emph{Proceedings of the AAAI Conference on Artificial Intelligence}, 35\penalty0 (1):\penalty0 12--20, May 2021.
\newblock ISSN 2374-3468, 2159-5399.
\newblock \doi{10.1609/aaai.v35i1.16072}.
\newblock URL \url{https://ojs.aaai.org/index.php/AAAI/article/view/16072}.

\bibitem[Geng et~al.(2023)Geng, Ziqing, Chihsu, and Jiamin]{CGPM_2023}
Zhao Geng, Gao Ziqing, Tsai Chihsu, and Lu~Jiamin.
\newblock Cgpm: Poverty mapping framework based on multi-modal geographic knowledge integration and macroscopic social network mining.
\newblock In Massih-Reza Amini, St{\'e}phane Canu, Asja Fischer, Tias Guns, Petra Kralj~Novak, and Grigorios Tsoumakas, editors, \emph{Machine Learning and Knowledge Discovery in Databases}, pages 549--564, Cham, 2023. Springer Nature Switzerland.
\newblock ISBN 978-3-031-26419-1.

\bibitem[Akbari et~al.(2021)Akbari, Yuan, Qian, Chuang, Chang, Cui, and Gong]{akbari_vatt:_2021}
Hassan Akbari, Liangzhe Yuan, Rui Qian, {Wei-Hong} Chuang, {Shih-Fu} Chang, Yin Cui, and Boqing Gong.
\newblock {VATT} : Transformers for multimodal self - supervised learning from raw video , audio and text, December 2021.
\newblock URL \url{http://arxiv.org/abs/2104.11178}.
\newblock arXiv:2104.11178 [cs, eess].

\bibitem[Ratledge et~al.(2022)Ratledge, Cadamuro, De~La~Cuesta, Stigler, and Burke]{ratledge_using_2022}
Nathan Ratledge, Gabe Cadamuro, Brandon De~La~Cuesta, Matthieu Stigler, and Marshall Burke.
\newblock Using machine learning to assess the livelihood impact of electricity access.
\newblock \emph{Nature}, 611\penalty0 (7936):\penalty0 491--495, November 2022.
\newblock ISSN 0028-0836, 1476-4687.
\newblock \doi{10.1038/s41586-022-05322-8}.
\newblock URL \url{https://www.nature.com/articles/s41586-022-05322-8}.

\bibitem[Park et~al.(2022)Park, Han, Ahn, Kim, Yang, Lee, Hong, Kim, Park, Yang, and Cha]{park_learning_2022}
Sungwon Park, Sungwon Han, Donghyun Ahn, Jaeyeon Kim, Jeasurk Yang, Susang Lee, Seunghoon Hong, Jihee Kim, Sangyoon Park, Hyunjoo Yang, and Meeyoung Cha.
\newblock Learning economic indicators by aggregating multi-level geospatial information.
\newblock \emph{Proceedings of the AAAI Conference on Artificial Intelligence}, 36\penalty0 (11):\penalty0 12053--12061, June 2022.
\newblock ISSN 2374-3468, 2159-5399.
\newblock \doi{10.1609/aaai.v36i11.21464}.
\newblock URL \url{https://ojs.aaai.org/index.php/AAAI/article/view/21464}.

\bibitem[Engstrom et~al.(2021)Engstrom, Hersh, and Newhouse]{engstrom_poverty_2021}
Ryan Engstrom, Jonathan Hersh, and David Newhouse.
\newblock \emph{Poverty from Space : Using High Resolution Satellite Imagery for Estimating Economic Well -being}.
\newblock Published by Oxford University Press on behalf of the World Bank, July 2021.
\newblock \doi{10.1596/40907}.
\newblock URL \url{https://hdl.handle.net/10986/40907}.

\bibitem[Doll et~al.(2006)Doll, Muller, and Morley]{doll_mapping_2006}
Christopher~N.H. Doll, {Jan-Peter} Muller, and Jeremy~G. Morley.
\newblock Mapping regional economic activity from night-time light satellite imagery.
\newblock \emph{Ecological Economics}, 57\penalty0 (1):\penalty0 75--92, April 2006.
\newblock ISSN 09218009.
\newblock \doi{10.1016/j.ecolecon.2005.03.007}.
\newblock URL \url{https://linkinghub.elsevier.com/retrieve/pii/S0921800905001254}.

\bibitem[Sutton and Costanza(2002)]{sutton_global_2002}
Paul~C. Sutton and Robert Costanza.
\newblock Global estimates of market and non-market values derived from nighttime satellite imagery, land cover, and ecosystem service valuation.
\newblock \emph{Ecological Economics}, 41\penalty0 (3):\penalty0 509--527, June 2002.
\newblock ISSN 09218009.
\newblock \doi{10.1016/S0921-8009(02)00097-6}.
\newblock URL \url{https://linkinghub.elsevier.com/retrieve/pii/S0921800902000976}.

\bibitem[{Pérez-Sindín} et~al.(2021){Pérez-Sindín}, Chen, and Prishchepov]{PEREZSINDIN2021100647}
{Xaquín}~S. {Pérez-Sindín}, {Tzu-Hsin}~Karen Chen, and Alexander~V. Prishchepov.
\newblock Are night-time lights a good proxy of economic activity in rural areas in middle and low-income countries? examining the empirical evidence from colombia.
\newblock \emph{Remote Sensing Applications: Society and Environment}, 24:\penalty0 100647, 2021.
\newblock ISSN 2352-9385.
\newblock \doi{https://doi.org/10.1016/j.rsase.2021.100647}.
\newblock URL \url{https://www.sciencedirect.com/science/article/pii/S235293852100183X}.

\bibitem[Liu et~al.(2021)Liu, He, Bai, Liu, Wu, Zhao, and Yang]{liu_erratum:_2021}
Haoyu Liu, Xianwen He, Yanbing Bai, Xing Liu, Yilin Wu, Yanyun Zhao, and Hanfang Yang.
\newblock Nightlight as a proxy of economic indicators : Fine - grained gdp inference around chinese mainland via attention - augmented cnn from daytime satellite imagery.
\newblock \emph{Remote Sensing}, 13\penalty0 (17):\penalty0 3360, August 2021.
\newblock ISSN 2072-4292.
\newblock \doi{10.3390/rs13173360}.
\newblock URL \url{https://www.mdpi.com/2072-4292/13/17/3360}.

\bibitem[Proville et~al.(2017)Proville, {Zavala-Araiza}, and Wagner]{proville_night-time_2017}
Jeremy Proville, Daniel {Zavala-Araiza}, and Gernot Wagner.
\newblock Night-time lights: A global, long term look at links to socio-economic trends.
\newblock \emph{PLOS ONE}, 12\penalty0 (3):\penalty0 e0174610, March 2017.
\newblock ISSN 1932-6203.
\newblock \doi{10.1371/journal.pone.0174610}.
\newblock URL \url{https://dx.plos.org/10.1371/journal.pone.0174610}.

\bibitem[Gibson et~al.(2021)Gibson, Olivia, {Boe-Gibson}, and Li]{gibson_which_2021}
John Gibson, Susan Olivia, Geua {Boe-Gibson}, and Chao Li.
\newblock Which night lights data should we use in economics, and where?
\newblock \emph{Journal of Development Economics}, 149:\penalty0 102602, March 2021.
\newblock ISSN 03043878.
\newblock \doi{10.1016/j.jdeveco.2020.102602}.
\newblock URL \url{https://linkinghub.elsevier.com/retrieve/pii/S0304387820301772}.

\bibitem[Hassan et~al.(2011)Hassan, Sanchez, and Yu]{hassan_financial_2011}
M.~Kabir Hassan, Benito Sanchez, and {Jung-Suk} Yu.
\newblock Financial development and economic growth: New evidence from panel data.
\newblock \emph{The Quarterly Review of Economics and Finance}, 51\penalty0 (1):\penalty0 88--104, February 2011.
\newblock ISSN 10629769.
\newblock \doi{10.1016/j.qref.2010.09.001}.
\newblock URL \url{https://linkinghub.elsevier.com/retrieve/pii/S1062976910000645}.

\bibitem[Horvath et~al.(2021)Horvath, Baireddy, Hao, Montserrat, and Delp]{horvath_manipulation_2021}
Janos Horvath, Sriram Baireddy, Hanxiang Hao, Daniel~Mas Montserrat, and Edward~J. Delp.
\newblock Manipulation detection in satellite images using vision transformer.
\newblock In \emph{2021 {IEEE}/{CVF} {Conference} on {Computer} {Vision} and {Pattern} {Recognition} {Workshops} ({CVPRW})}, pages 1032--1041, Nashville, TN, USA, June 2021. IEEE.
\newblock ISBN 9781665448994.
\newblock \doi{10.1109/CVPRW53098.2021.00114}.
\newblock URL \url{https://ieeexplore.ieee.org/document/9522892/}.

\bibitem[Chen et~al.(2021)Chen, Fan, and Panda]{chen_crossvit:_2021}
{Chun-Fu}~Richard Chen, Quanfu Fan, and Rameswar Panda.
\newblock {CrossViT} : Cross - attention multi - scale vision transformer for image classification.
\newblock In \emph{2021 {IEEE}/{CVF} {International} {Conference} on {Computer} {Vision} ({ICCV})}, pages 347--356, Montreal, QC, Canada, October 2021. IEEE.
\newblock ISBN 9781665428125.
\newblock \doi{10.1109/ICCV48922.2021.00041}.
\newblock URL \url{https://ieeexplore.ieee.org/document/9711309/}.

\bibitem[Han and Zhang(2015)]{han_spark:_2015}
Zhijie Han and Yujie Zhang.
\newblock Spark: a big data processing platform based on memory computing.
\newblock In \emph{2015 {Seventh} {International} {Symposium} on {Parallel} {Architectures}, {Algorithms} and {Programming} ({PAAP})}, pages 172--176, Nanjing, China, December 2015. IEEE.
\newblock ISBN 9781467391177.
\newblock \doi{10.1109/PAAP.2015.41}.
\newblock URL \url{http://ieeexplore.ieee.org/document/7387321/}.

\end{thebibliography}

\end{document}